\documentclass[conference]{IEEEtran}
\usepackage{fancyhdr}
\usepackage{graphicx}
\usepackage{bm}
\usepackage[linesnumbered,boxed]{algorithm2e}
\usepackage{fixltx2e}
\usepackage{amsmath}

\usepackage{caption}
\usepackage[font=footnotesize]{subfig}

\makeatletter
\newcommand{\removelatexerror}{\let\@latex@error\@gobble}
\makeatother

\begin{document}
%
\title{Automatic Interpretation of Unordered Point Cloud Data for UAV Navigation in Construction}

\author{\IEEEauthorblockN{M.D. Phung, C.H. Quach, D.T. Chu}
\IEEEauthorblockA{University of Engineering and Technology\\Vietnam National University, Hanoi\\
Hanoi, Vietnam\\
Email: duongpm@vnu.edu.vn}
\and
\IEEEauthorblockN{N.Q. Nguyen}
\IEEEauthorblockA{National Institute of Information\\ and Communications Strategy\\
Hanoi, Vietnam}
\and
\IEEEauthorblockN{T.H. Dinh and Q.P. Ha}
\IEEEauthorblockA{School of Electrical, Mechanical\\ and Mechatronic Systems\\
University of Technology Sydney\\
Sydney, Australia\\
Quang.Ha@uts.edu.au}}

\maketitle

\thispagestyle{fancy}
\fancyhead{}
\lhead{}
\lfoot{}
\cfoot{}
\rfoot{}
\renewcommand{\headrulewidth}{0pt}
\renewcommand{\footrulewidth}{0pt}

\begin{abstract}
The objective of this work is to develop a data processing system that can automatically generate waypoints for navigation of an unmanned aerial vehicle (UAV) to inspect surfaces of structures like buildings and bridges. The input includes data recorded by two 2D laser scanners, orthogonally mounted on the UAV, and an inertial measurement unit (IMU). To achieve the goal, algorithms are developed to process the data collected. They are separated into three major groups: (i) the data registration and filtering to generate a 3D model of the structure and control the density of point clouds for data completeness enhancement; (ii) the surface and obstacle detection to assist the UAV in monitoring tasks; and (iii) the waypoint generation to set the flight path. Experiments on different data sets show that the developed system is able to reconstruct a 3D point cloud of the structure, extract its surfaces and objects, and generate waypoints for the UAV to accomplish inspection tasks.
\end{abstract}

\IEEEpeerreviewmaketitle

\section{Introduction}
Unmanned aerial vehicles (UAVs) are expected to yield automatic solutions for inspecting and monitoring large and hardly accessible structures like bridges, towers, dams, wind turbines and heritage monuments due to their flexibility in operating space and ability to carry specialized sensory equipment. In \cite{Eschmann2012}, a micro air vehicle system was employed to scan buildings using a high resolution camera. Pictures taken were successfully stitched with sufficient quality for crack and damage detection. In \cite{Hallermann2013}, an advanced UAV system were developed to evaluate the state of historical monuments including a chimney built in early 1980\textsuperscript{th} and a natural stone masonry tower inclined $4.7^0$. Images captured after processing revealed damages in both monuments. A control system for navigation of the UAV from an initial to a final position in an unknown 3D environment was used to monitor and maintain bridges \cite{Metni2007}. UAVs were also used to inspect and monitor oil-gas pipelines, roads, power generation grids and other essential infrastructure \cite{Rathinam2008}.

In order to complete those inspection missions, the UAV system typically carries out three steps including creating waypoints from coarse data of the environment, controlling the UAV to follow waypoints and collect data of the structure, and processing the data to detect defects or damages. Among these steps, the generation of waypoints is still manual with little support of automation. In \cite{Eschmann2012}, the UAV was directly controlled by a pilot. In \cite{Hallermann2013}, the flight plan was set by the user while work \cite{Metni2007} focused on a control strategy without concerning obstacles. As the result, the deployment process is time consuming and the data collected are often redundant and ineffective. For example, among more than 12,000 images taken in \cite{Hallermann2013} only several hundred images were finally valid for stitching and inspection.

In this study, we introduce an automatic solution to generate waypoints for the UAV to inspect and monitor built infrastructure. Input data includes range information acquired by two laser scanners and orientation measured by an IMU. The system then can align the range data into point clouds and register them to represent a 3D model of the structure. Based on the model, the system detects surfaces, extracts borders and clusters obstacle objects. For each surface to be inspected, the system will generate waypoints which are optimized for path planning and obstacle avoidance. The total processing time, from loading data to generating waypoint, is several minutes for the surface with area of hundreds of squared meters.

\section{Methodology}
\begin{figure*}[!t]
	\centering
	\includegraphics[width=6in]{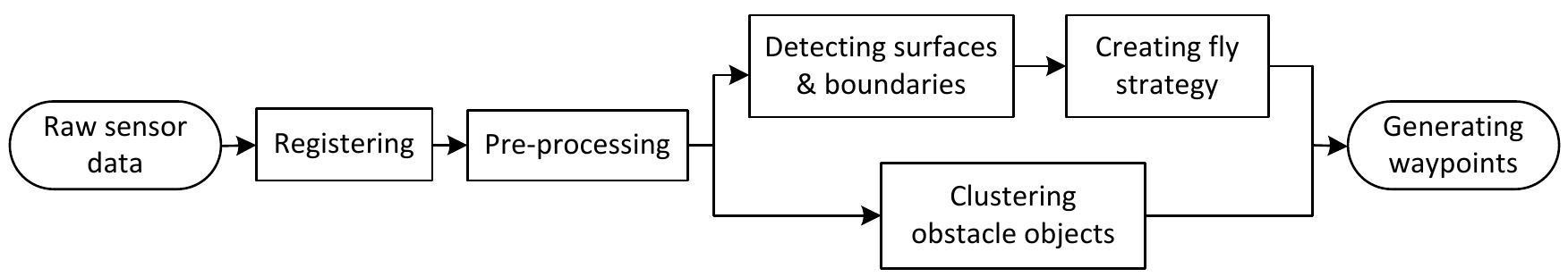}
	\caption{An overview of the system.}
	\label{fig:systemflowchart}
\end{figure*}

Figure \ref{fig:systemflowchart} presents an overview of the system, consisting of seven stages. Details of each stage are explained as follows.

\subsection{Data collection}
\begin{figure}[!t]
		\centering
		\includegraphics[scale=0.25]{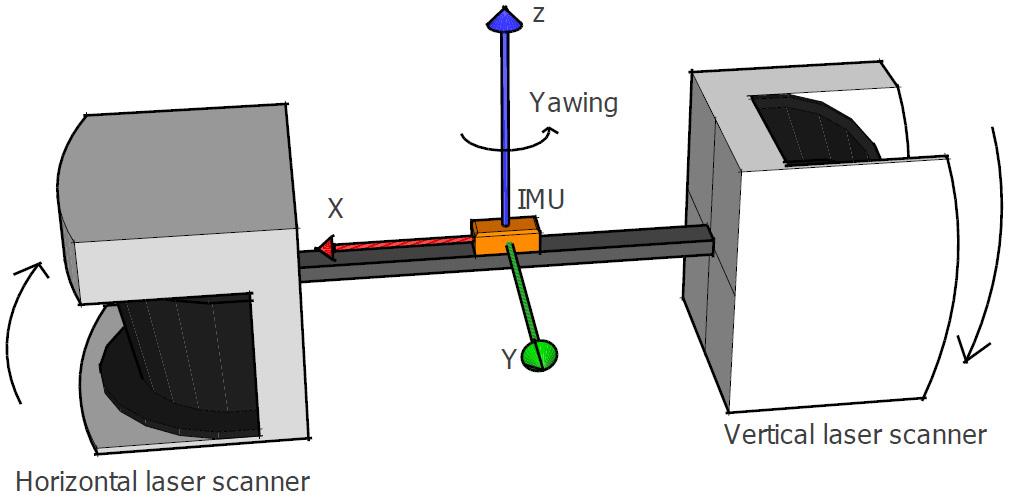}
		\caption{Hardware configuration for data collection.}
		\label{fig:datacollection}
	\end{figure}

Data used for the system includes range information measured by laser scanners. Due to the limitation in payload and energy capability, the 3D laser scanner is not suitable for the UAV. Instead, we use two 2D laser scanners and an IMU to collect information of the structure. The arrangement is shown in Fig.~\ref{fig:datacollection}. The two laser scanners are placed orthogonally to each other and symmetrically with respect to the center of the UAV. One scans horizontally and the other scans vertically. The IMU is fixed at the center of the UAV. The data collection is carried out by yawing the UAV $360^\circ$. During yawing, points acquired by vertical scanning will be used to generate a 3D point cloud of the structure.

Let $(\rho_i, \alpha_i)$ be the distance and angle of point $p_i$ measured by the vertical scanner. Its Cartesian coordinates in the local frame attached to the center of the scanner are given by:
\begin{IEEEeqnarray}{lCr}
x^*_i=-\rho_icos(\alpha_i) \nonumber \\
y^*_i=0 \\
z^*_i=-\rho_isin(\alpha_i) \nonumber.
\end{IEEEeqnarray}
As the scanner moves during acquiring data, it is required to map local coordinates to a fixed global frame. Let the origin of the global frame be the center of UAV at the initial scan, associated with $x-$,$y-$,and $z-$ coordinates as shown in Fig.~\ref{fig:datacollection}. The laser scanner motion is then the combination of two components: rotation caused by yawing and translation caused by the external forces like wind acting on the UAV. Let $R$ and $T$ be respectively the rotation and translation matrices of the motion. The coordinate vector of point $p_i$ in the global frame, $\bm{x}_i=[x_i \: y_i \: z_i]^T$, is given by:
\begin{eqnarray}
\bm{x}_i=R_i\bm{x}_i^* + T_i.
\end{eqnarray}

\begin{figure}[!t]
\centering
\subfloat[]{ \label{fig:ICPa} \includegraphics[width=1.8in]{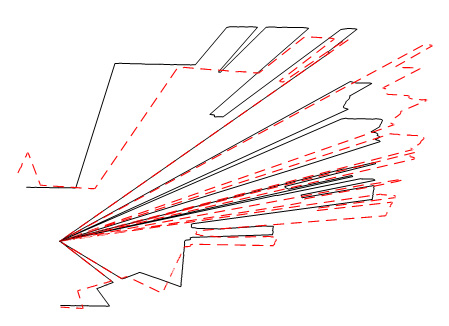}}
\subfloat[]{ \label{fig:ICPb} \includegraphics[width=1.6in]{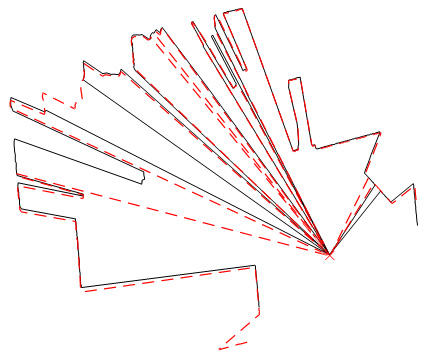}}
\centering
\caption{Horizontal scans: (a) before applying the ICP algorithm, and (b) after applying the ICP algorithm.}
\label{fig:icp}
\end{figure}

In our system, $R_i$ is directly measured by the IMU while $T$ is determined by the horizontal scanner as follows: During yawing, the horizontal scanner will scan the same surface of the structure with small changes in position and orientation. The scans therefore have much in common (Fig.~\ref{fig:ICPa}) so that we can use the iterative closest point (ICP) algorithm \cite{Besl1992} to match them and extract the translation and rotation (Fig.~\ref{fig:ICPb}). As the rotation is already known, we incorporate it by rotating the raw scans before applying the ICP algorithm. This customization can improve the accuracy of the estimation and reduce the computation load.

\subsection{Point cloud registration}
After collecting data, point clouds recorded at different positions need to be combined to a complete 3D model of the structure. This process is called registration. The key idea is to identify corresponding points between the point clouds and to find a transformation that minimizes the distance between them. We carry it out by first predicting the overlaps between point clouds based on their recorded positions. We then apply the ICP algorithm to the predicted overlaps to obtain the transformation.

\subsection{Pre-processing: point cloud filtering and voxelization}
Due to characteristics of the laser scanner, the point cloud generated in the registration process typically contains sparse outliers and varies in densities. This complicates the estimation of point cloud features such as surface normals or curvature changes, leading to erroneous values, which, in turn, might cause failure in further processing steps. The objective of the pre-processing process is hence to filter sparse outliers and equalize point densities.

The sparse outlier removal is carried out based on computation of the Gaussian distribution of distances from each point to its neighbors. By comparing the mean, $\mu$, and standard deviation, $\sigma$,  with pre-defined global thresholds, a point could be considered as an outlier and trimmed from the dataset if it falls outside the $\mu\pm d_t\sigma$, where $d_t$ is a threshold coefficient \cite{Rusu2008}. For density equalization, the voxelization technique is employed. A uniformly spaced 3D voxel grid is created over the input point cloud. All points within a voxel are then presented by the centroid of the voxel. This process can largely reduce the amount of points, but may introduce a small amount of geometric error due to quantization.

\begin{figure}
\subfloat{
			\includegraphics[width=115 pt]{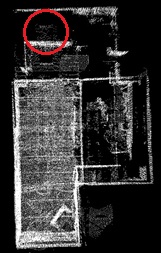}
}
\subfloat{
	\includegraphics[width=110 pt]{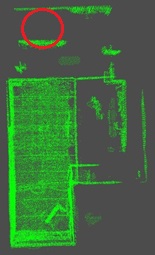}
}
		\caption{Point clouds before and after filtering and voxelizing.}
		\label{fig:filter}
\end{figure}

Fig.~\ref{fig:filter} demonstrates an example of filtering and applying voxelization on a dataset. Here, the number of scan points in the original cloud was cut down from 90,396 points (left) to 80,114 points (right) while isolated points were considered as outliers and removed.

\begin{figure}
	\centering
	\includegraphics[width=140 pt]{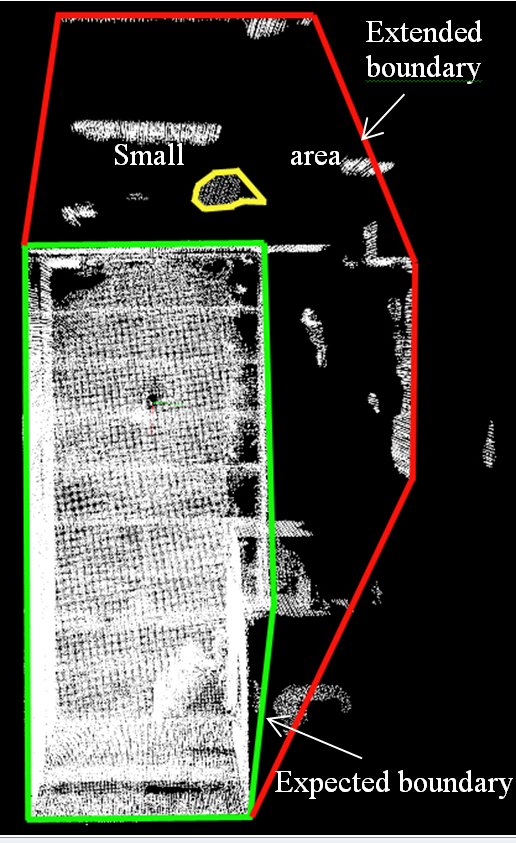}
	\caption{Problems related to surface and boundary detection.}
	\label{fig:surface_detection}
\end{figure}

\subsection{Surface and boundary detection}
Surfaces are the main target to be inspected so they need to be extracted from the point cloud. We use the RANSAC (random sample consensus) method and planar model for this task due their simplicity and robustness \cite{Fischler1981}. Since a plane model $(ax + by + cz +d = 0)$ is given, $M=[a,b,c,d]^T$ is the parameter vector to be identified. The estimate procedure is a repeat of the following steps:

\begin{itemize}
	\item First, a random subset of the original data is selected.
	\item A planar model is then fitted to the selected subset.
	\item Remaining points are then tested against the model. A point is considered as an inlier if its distance to the model is smaller than a pre-defined threshold. The fitting model is reasonable good if it has sufficient inliers.
	\item The fitting model is refined to better fit all found inliers.
\end{itemize}
In our system, a threshold value of 20 cm and a repeat of 200 times give compromised performance.

After detecting surfaces, their boundaries need to be determined. For each surface, corresponding inliers are first projected on the found planar model. The 2D convex hull algorithm \cite{Barber1996} is then employed to fit a bounding polygon to the projected inliers.

\begin{figure}
	\centering
	\includegraphics[width=3 in]{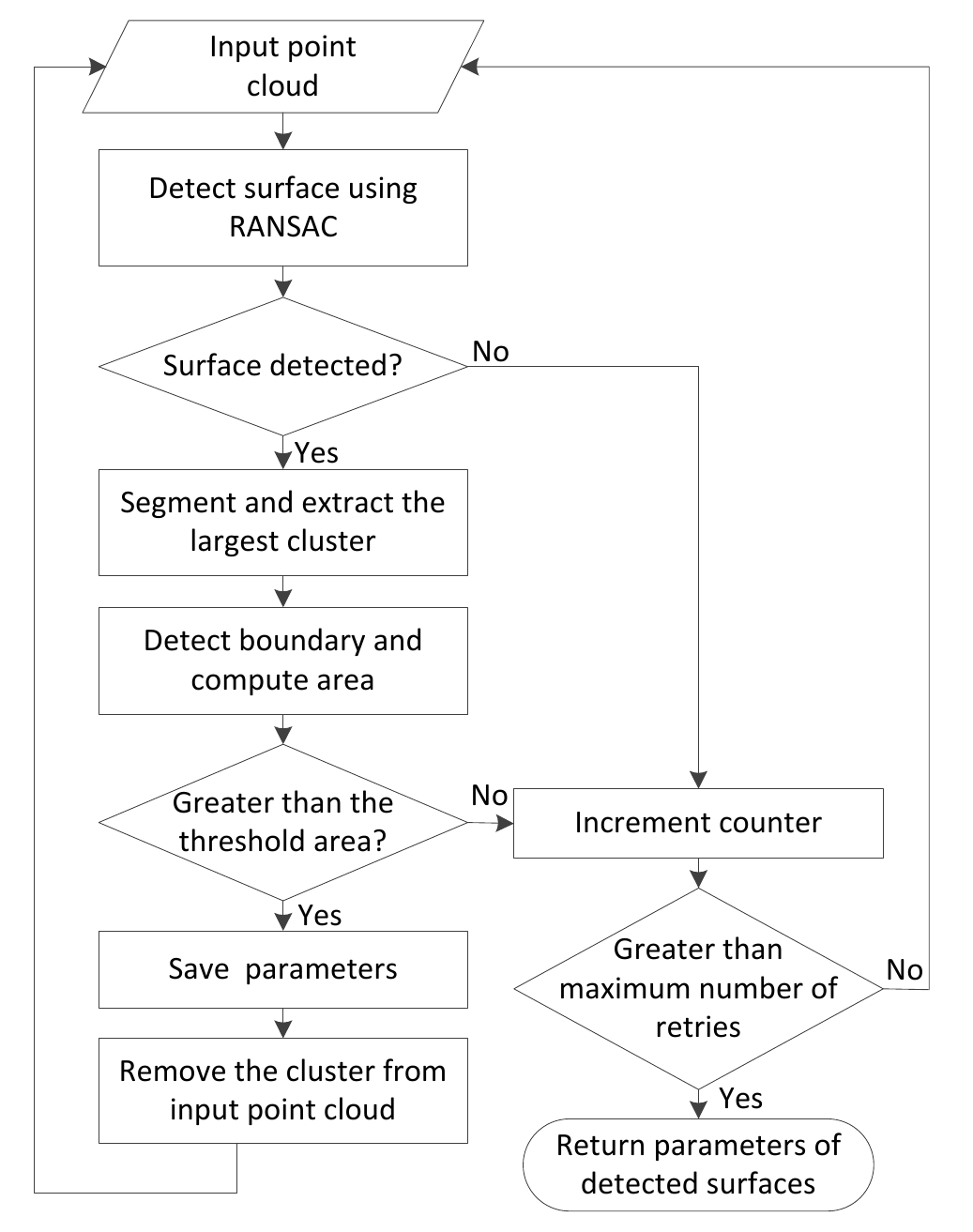}
	\caption{Flowchart of the surface and boundary detection.}
	\label{fig:algorithm}
\end{figure}

In practice, two problems may arise during using the RANSAC and convex hull algorithms as illustrated in Fig.~\ref{fig:surface_detection}. First, the detected surface may contain isolated clusters causing the boundary to be extended. Second, when detecting small surfaces, the RANSAC may cause confusion between surfaces belonging to the structure and objects. In order to deal with the first problem, we require a high point density of a surface. We therefore trim out the isolated groups by clustering the surface and only keep the largest one. The clustering algorithm will be presented in Section \ref{clustering}. For the second problem, the inlier quantity and density are used to compute the area of each detected surface, which is limited to a threshold set by the user. Fig.~\ref{fig:algorithm} shows a flowchart of the surface and boundary detection.

\subsection{Obstacle clusterization}
\label{clustering}
The scanned points that do not belong to any surfaces maintain their importance in UAV navigation. It is suggested here to cluster them into small groups as obstacles in order to support the path finding algorithm and UAV's operator to have a better vision of the environment. The clusterization is carried out by finding the nearest neighbour that is essentially similar to a flood fill algorithm \cite{Rusu2009}, as shown in Fig.~\ref{fig:pseudocode}. We use the 3D Kd-tree structure to speed up the neighbour searching process. We then use the octree structure to represent the clusters for visualization.

\begin{figure}
	\removelatexerror
	{\fontsize{08}{08}\selectfont
		\begin{algorithm*} [H]
			\SetAlgoLined
			\KwData{A point cloud $P$}
			\KwResult{A list of clusters $L$}
			Init an empty queue $Q$\;
			\For {each point $p_i \in P$} {
				\If{$p_i$ was not processed before} {
					Add it to $Q$\;
					Mark it as processed\;
					\For{each point $q_j \in Q$}{
						Search for its neighbors within a sphere of radius $ r < \epsilon$\;
						\For{each found neighbor $q_j^k$}{
							\If{$q_j^k$ was not processed before}{
								Add it to $Q$\;
								Mark it as processed\;
							}
						}
					}
					Add $Q$ to the end of $L$\;
					Reset $Q$\;
				}
			}
		\end{algorithm*}
	}
	\caption{Pseudo code of the clustering algorithm.}
	\label{fig:pseudocode}
\end{figure}

\subsection{Flight strategy creation}
\begin{figure}
	\centering
	\includegraphics[width=170 pt]{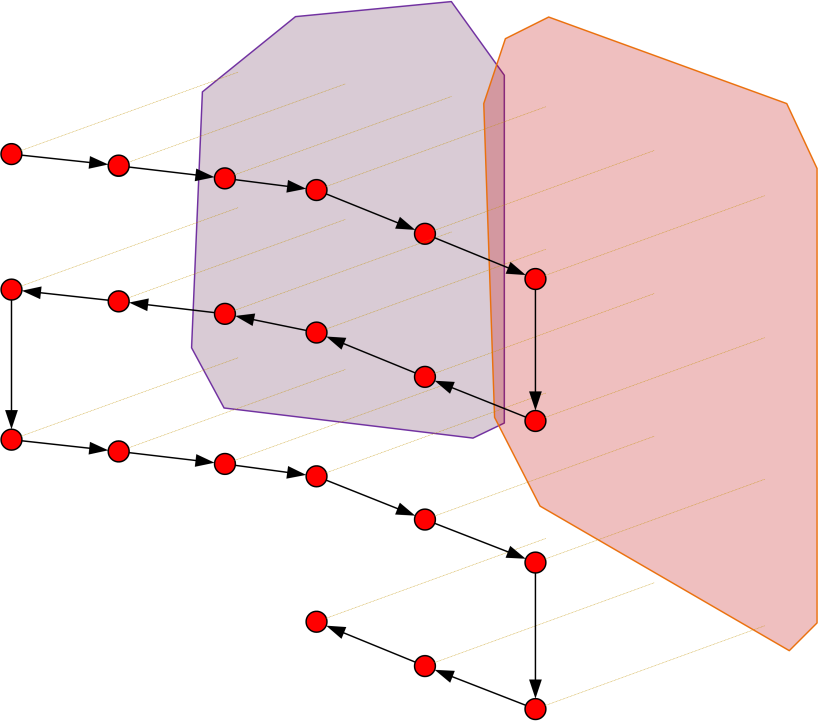}
	\caption{A flight strategy to take photos of surfaces for inspection.}
	\label{fig:flightstrategy}
\end{figure}

A flight strategy is a set of positions at which the UAV has to pause to collect data. In order to create a flight strategy, it is required to know the assigned tasks, requirements on data collected, and capability of the sensory devices. For instance, a common situation would be taking photos of a structure and sticking them together for defect or damage detection. In this case, requirements may include the minimum resolution of photos, at which defects are still detectable and the overlap percentage between consecutive photos feasible for sticking and mosaicking. For this, the capability of a sensory device includes the resolution of the camera, its open angle and the number of degrees of freedom of the mechanism on which the camera is mounted. Given those requirements and parameters, we can compute the area that each photo captures, positions to take photo, their order, number of positions needed to cover the surface and thus the flight strategy, as illustrated in Fig.~\ref{fig:flightstrategy}.

\begin{figure*}
	\subfloat[]{
		\label{fig:hypothesis_plane}
		\includegraphics[width=3in]{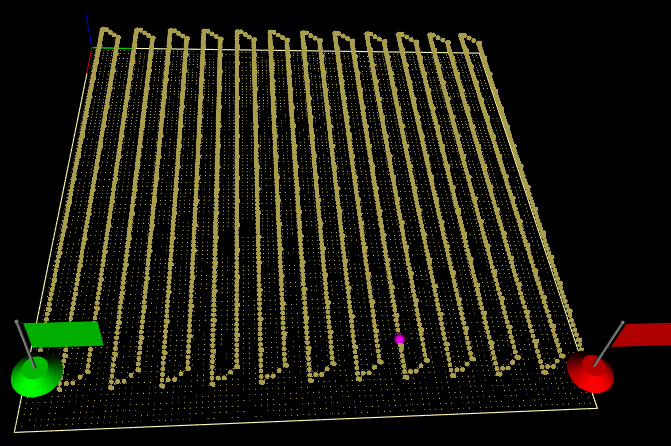}
	}
	\subfloat[]{	
		\label{fig:hypothesis_cube}
		\includegraphics[width=2.03in]{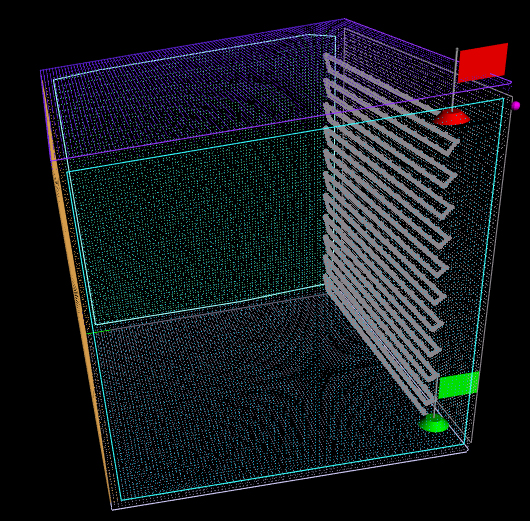}
	}
	\subfloat[]{
		\label{fig:hypothesis_cross}
		\includegraphics[width=2.05in]{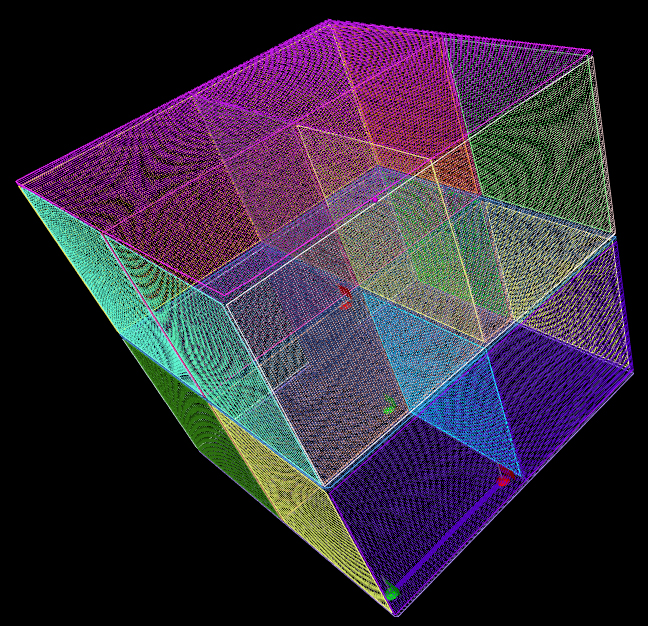}
	}	
	\centering
	\caption{Experiments on hypothesis data sets: (a) a single surface, (b) a single cube, and (c) a cube with crossed planes.}
\end{figure*}

\begin{figure*}
	\subfloat[]{
		\label{fig:brigde_a}
		\includegraphics[width=2.45in]{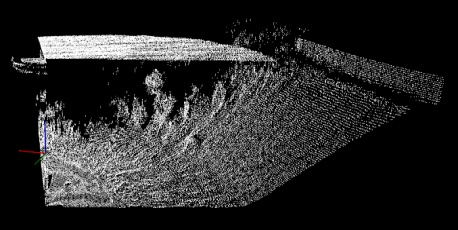}
	}
	\subfloat[]{
		\label{fig:brigde_b}
		\includegraphics[width=2.3in]{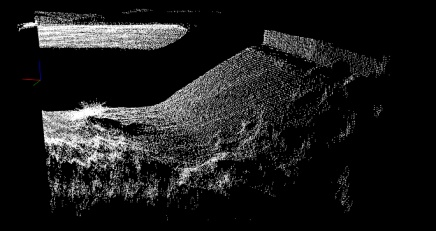}
	}
	\subfloat[]{
		\label{fig:brigde_c}
		\includegraphics[width=2.3in]{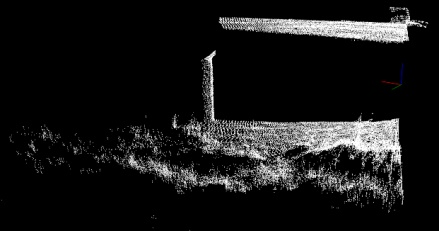}
	}\\
	
	\subfloat[]{
		\label{fig:brigde_d}
		\includegraphics[width=2.5in]{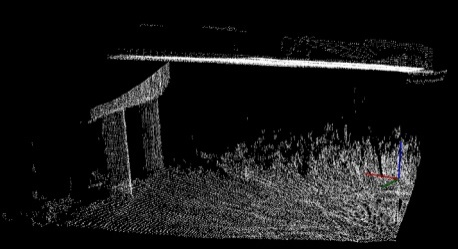}
	}
	\subfloat[]{
		\label{fig:brigde_all}
		\includegraphics[width=4.65in]{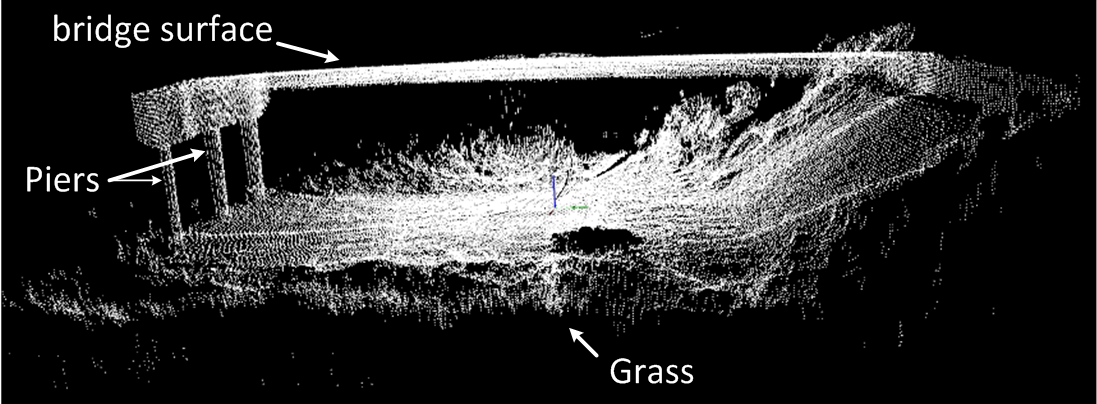}
	}					
	\caption{Point cloud registration: (a)-–(d) registered point clouds from raw data, and (e) merged point cloud.}
	\label{fig:registration_bridge}
\end{figure*}

\begin{figure*}
	\subfloat[]{
		\includegraphics[width=4.1in]{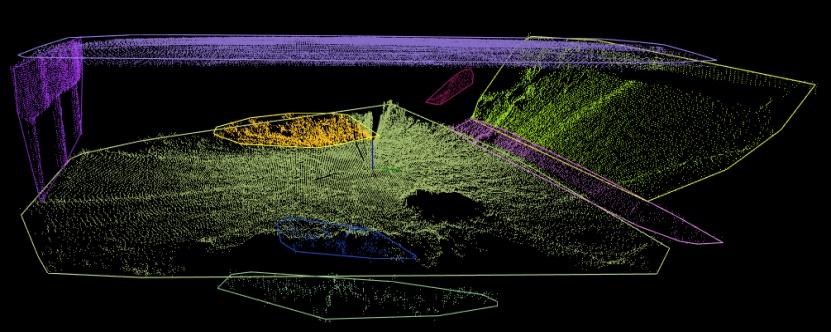}
	}
	\subfloat[]{
		\includegraphics[width=3.05in]{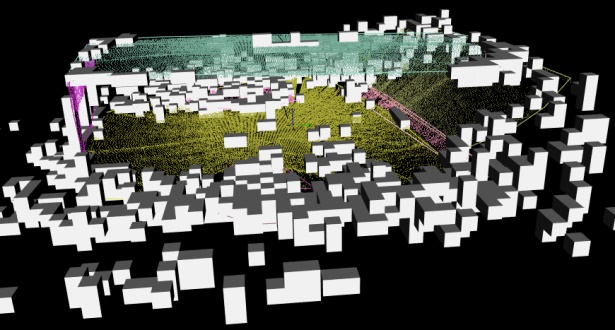}
	}
	\caption{ Surface and obstacle detection: (a) planar surfaces, and (b) obstacle objects.}
	\label{fig:plane_obstacle}	
\end{figure*}

\begin{figure*}
	\subfloat[]{
		\label{fig:waypoint1}
		\includegraphics[width=3.6in]{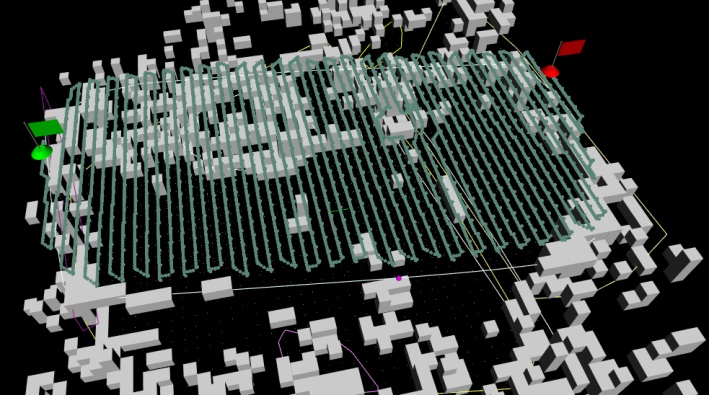}
	}
	\subfloat[]{
		\label{fig:waypoint2}
		\includegraphics[width=3.3in]{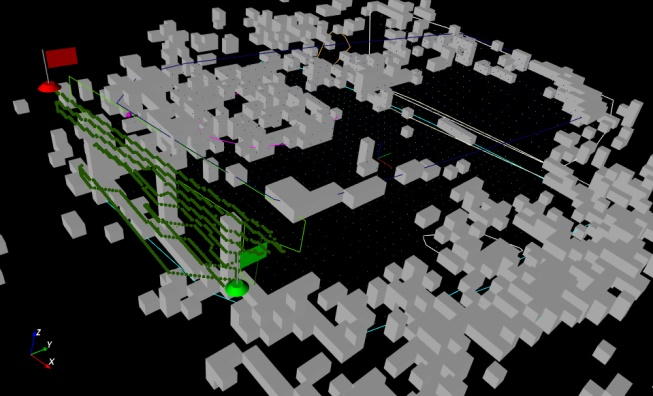}
	}
	\centering
	\caption{Waypoints generated to navigate the UV to take photos: (a) bridge surface, and (b) pier.}
\end{figure*}

\subsection{Waypoint generation}
The final stage in our system is the generation of waypoints. They are intermediate points the UAV follows to complete a flight strategy. Given the flight strategy, we can determine waypoints by using a path finding algorithm. We implement it by dividing the work space into a grid of voxels. Each voxel has the free or occupied status corresponding to the existence or not of object in that voxel. In order to consider the UAV as a particle moving without collision between voxels, we mark all free voxels in a sphere with the radius equal to the largest dimension of the UAV as occupied. We then use the A-star algorithm \cite{Hart1968} to find the shortest path between stop points. In each step, the cost to move from one voxel to another surrounding neighbor is computed as:
\begin{equation}
C(\alpha,\beta,\gamma)=a_1 \alpha^2+ a_2 \beta^2+a_3 \gamma^2, 
\end{equation}
where coordinates $\alpha,\beta,\gamma \in \{-1,0,1\}$  indicate the position of neighbor, and the coefficient $a_i$  assign a particular weight to each direction.

\section{Experiments}
The system was evaluated on various sets of data recording different structures like an office, a building or a bridge. The UAV used is a quadcopter with the size of 60 cm $\times$ 60 cm $\times$ 40 cm equipped with a camera mounted on a two degree-of-freedom gimbal, two laser scanners, an IMU and other electronic boards.

The laser scanner used is Hokuyo UTM-30LX with the scanning range of 30 m, detection angle of $270^\circ$, angular resolution of $0.25^\circ$ and scan period of 25 ms. The IMU is Xsense Mti-10 outputting the rotation matrix with the frequency of 2 KHz, accuracy of $0.5^\circ$, and no accumulated error. The software is written in C++ and run on a PC with the processor Intel Core2 Dual 2.53 GHz and 4 GB RAM.

\subsection{Experiments on hypothesis data sets}
Experiments are first carried out on hypothesis data sets to evaluate the behavior of system in regular and irregular cases. They include the point clouds containing a single point, a single line, a single surface, a single cube and a cube with crossed planes. For the two first cases, no surface is detected. For the two later cases, all surfaces and their boundaries are detected. 

The flight strategy and waypoints for each surface are also created as shown in Fig.~\ref{fig:hypothesis_plane} and Fig.~\ref{fig:hypothesis_cube}. For the last case, all surfaces and boundaries are detected, but waypoints are not generated because the surfaces are blocked by cross planes (Fig.~\ref{fig:hypothesis_cross}).

\subsection{Experiments on real data sets}
After evaluating the behavior of system on hypothesis data, experiments on real data sets are carried out. Fig.~\ref{fig:registration_bridge} shows results of aligning raw scans into four point clouds (Fig.~\ref{fig:brigde_a}--Fig.~\ref{fig:brigde_d}) and registering them to become a complete part of the bridge with the size of 22 m $\times$ 10 m $\times$ 4.5 m (Fig.~\ref{fig:brigde_all}). The alignment error, measured by yawing $360^\circ$ and comparing the first and last scans, is 10 cm. Fig.~\ref{fig:plane_obstacle} represents the planes and objects detected with the area threshold of 2 m$^2$. 

Besides planes created by grasses and trees, all main planes including the bridge surface, pier, sides and land surface are detected. The boundaries are reliable for the convex surfaces. However, for the concave surfaces like pier, extra areas should be accounted for. We deal with this problem by adding a function to allow manually modification of the boundary. In practice, this function is useful for not only the concave but also convex surfaces as it allows the operator to ensure the accuracy of detected boundaries before generating waypoints for inspection.

Figure \ref{fig:waypoint1} shows the flight strategy and waypoints generated for the task of taking photos of the bridge surface with the coverage of 60 cm $\times$ 40 cm for each photo and the overlap of $20\%$ between consecutive photos. A total of 1,146 stop points and 6,699 waypoints were generated to cover the area of 220 m$^2$. The path generated does not collide with any obstacles appearing in scans. The result is similar for the pier surface with 75 stopover points and 891 waypoint generated (Fig.~\ref{fig:waypoint2}). The processing time for a surface, from loading raw data to generating waypoints, averages 3 minutes in which $83\%$ is consumed on aligning and registering point clouds. This result is very encouraging because manually defining waypoints for a surface often takes a lot of time or even becomes impractical for large structures like kilometer-long bridges. The waypoint generation will be further improved in a future work involving the deployment of a group of UAVs \cite{Woods2016} in coordinated formation \cite{Nguyen2004} to enhance the scanning coverage.

\section{Conclusion}
In this study, we have introduced a paradigm to collect and process data for automatic navigation of UAVs in built infrastructure inspection. A hardware configuration was proposed to acquire sufficient information to reconstruct the 3D model of the structure while feasible to equip on a micro UAV. The stages necessary to process point cloud data and extract semantic information from them are introduced. At each stage, we have presented details of algorithms used and customization to enhance the performance. Through experiments, it is shown that the proposed system provides a quick and reliable solution for the automatic operation of the UAV used in built infrastructure inspection.

\section*{Acknowledgment}
The first author would like to acknowledge an Endeavours Research Fellowship by the Australian Government for the support (ERF\_PDR\_142403\_2015), in part, for this work.

\end{document}